\documentclass[letterpaper, 10 pt, conference]{ieeeconf}
\IEEEoverridecommandlockouts 
\usepackage{graphicx}
\usepackage{algorithm}
\usepackage{algorithmicx}
\usepackage[noend]{algpseudocode}
\usepackage{cite}
\usepackage{color}
\usepackage{amsmath,amssymb,amsfonts,dsfont}
\usepackage{bm}
\usepackage{setspace}
\usepackage{multirow}
\usepackage{mathtools}
\usepackage{fancyhdr}
\usepackage{authblk}
\usepackage{url}
\usepackage{tikz}
\usepackage{capt-of}
\usetikzlibrary{positioning,fit,calc,arrows}

\makeatletter
\def\BState{\State\hskip-\ALG@thistlm}
\makeatother

\usepackage[
  separate-uncertainty = true,
  multi-part-units = repeat
]{siunitx}

\title{\LARGE \bf
The Helping Hand: An Assistive Manipulation Framework Using Augmented
Reality and Tongue-Drive Interfaces}
\author{Fu-Jen Chu, Ruinian Xu, Zhenxuan Zhang, Patricio A. Vela, 
        and Maysam Ghovanloo
\thanks{Institute for Robotics and Intelligent Machines, 
        Georgia Institute of Technology, GA, USA
        {\tt\small \{fujenchu, gtsimonxu, james.zhang, pvela, mgh\}@gatech.edu}}%
\thanks{* This work was supported in part by NSF Award \#1605228.}%
}

\begin{document}
\maketitle
\thispagestyle{empty}
\pagestyle{empty}

\begin{abstract}


A human-in-the-loop system is proposed to enable collaborative
manipulation tasks for person with physical disabilities. Studies 
show that the cognitive burden of subject reduces with increased autonomy of
assistive system. Our framework obtains high-level intent from the user to
specify manipulation tasks. The system processes sensor input to interpret
the user's environment. Augmented reality glasses provide ego-centric
visual feedback of the interpretation and summarize robot affordances on a
menu. A tongue drive system serves as the input modality for triggering a
robotic arm to execute the tasks. Assistance experiments compare the system
to Cartesian control and to state-of-the-art approaches. Our system achieves
competitive results with faster completion time by simplifying manipulation
tasks.

	
\end{abstract}

\section{Introduction}
Paralysis afflicts 5.5 million people in the United States
\cite{Christopher2011}.  Persons with high-level paralysis rely on
caregivers and/or environmental modifications to accomplish the
activities of daily living (ADL). 
While the adoption of personal mobility devices and environmental
control systems provide some autonomy \cite{DiEtAl_AT[2003],huo2010evaluation},
there is still a gap between the activities enabled by these
interventions and the needs of the paralyzed population regarding the
ADL.  Research and translational efforts in robotics and assistive
technologies (AT) indicate that these support technologies can bridge
the existing ability gap. 




Assistive robotic manipulators have long been considered as enabling
technologies for self-supportiveness and independence in accomplishing
ADLs \cite{ChEtAl_RAM[2013],DeEtAl_AR[2008],GrHaSc_AR[2004]}.
Commonly seen assistive robotic arms such as the JACO arm
and the MANUS have 6-7 degrees of freedom, and admit execution of many
ADLs \cite{Jaco,KiWaBe_ToM[2012]}.
However, it is challenging for people with paralysis of arms
to fully control an assistive system at 
the required proficiency level
\cite{struijk2017wireless, LoStLo_ICORR[2017]}.
Even for non-paralyzed populations, the traditional manipulator control
interfaces require some level of expertise and 
exhibit occasional operator error \cite{KeSaCh_HRI[2017]}. Better performance can be achieved by
increasing robot autonomy 
\cite{ChEtAl_RAM[2013], ChWaCo_JSCM[2013]}.



The role of an assistive manipulator is to interact with the local
environment according to the desires of its user.
For persons with high-level paralysis, there is a need to
develop effective user interfaces for communicating human intent to the
robotic manipulator in a hands-free manner.
The Tongue Drive System (TDS) is a wireless assistive technology for
translating tongue motion to discrete commands \cite{huo2008magneto,
huo2010evaluation, kim2013tongue}. 
Studies show that it is an
effective hands-free interface, with high throughput and accuracy
as compared to other devices such as EEG, EMG, eye tracker, and Sip-and-Puff. 
TDS requires shorter training and calibration times (below 5 minutes);
users learn to interface the TDS quickly. Importantly, the tongue muscle
has a low rate of perceived exertion and does not fatigue easily. 


Meanwhile, the affordances of the robot assistant should be
communicated to the user in a seamless manner, so that they may select
what action to execute.  Visual display devices with dynamic menuing
provide the necessary flexibility, and are compatible with the TDS
interface.
Candidate display devices include laptops, tablets, audio assistants,
and augmented reality (AR) glasses
\cite{LaEtAl_APMR[2009]}.
Recently, AR has been applied to the rehabilitation and assistive
systems fields. With AR glasses, a user can control a virtual menu or
program \cite{azuma1997survey}.
Explored use cases include 
education for cognitively impaired school children \cite{richard2007augmented}, 
surgical robotics \cite{wen2014hand}, 
and prosthetic grasping assessment \cite{markovic2014stereovision}.

The recent studies \cite{wang2017robotic,weisz2017assistive} are most related to this work.
They use hands-free
interfaces (eye gaze and EEG, or sEMG) as input modalities to an assistive
robotic manipulator for performing pick and place, and grasp
planning activities.
Intermediate phases of the routine must be controlled by the user
through the assistance of a nearby monitor that provides AR feedback.
The AR approach was shown to improve task performance (time and error)
relative to the lack of AR.




\paragraph{Contribution}
Compared with prior approaches, our system's input modality is a
TDS and the visual display is a head-mounted AR system. 
The TDS is a robust interface for signalling intent with minimum burden 
even in noisy environments, making it more practical than other
interfaces.  
A headworn AR system, through head fixation, prevents gaze to be broken
from objects of interest, provides flexibility without an extra monitor,
and improves robot guidance by providing a virtual menu with possible
robot affordances. 
Our work improves the autonomous capabilities of the robotic arm through the
integration of modern computer vision algorithms and robotic planning
methods.  The overall system detects manipulable objects on nearby surfaces
and provides an AR menu interface for choosing to interact with
them.  The user selected high level menu commands signal intent to the
robotic arm, simplify the act of of manipulation for user desired tasks, and
lead to faster interaction times.



\section{System Architecture}
\begin{figure*}[t]
  \begin{tikzpicture}[outer sep=0pt]
  \definecolor{cobalt}{rgb}{0.0, 0.28, 0.67}
  \tikzstyle{outerBox}=[rounded corners=4pt,draw=red,thick,minimum height=0.75in];
  \tikzstyle{innerBox}=[rounded corners=2pt,draw=cobalt,thick, fill=gray!8, minimum width=1in];
  \tikzstyle{detailBox}=[rounded corners=2pt,draw=cobalt,thick, fill=gray!8,minimum width=1in,minimum height=5ex];

    \node[innerBox] (VS) at (0in,0in) {\bf Vision System};
    \node[detailBox,above of=VS]  (AVS) {\parbox{1in}{\centering \footnotesize
        Object Detection \\ 3D BBox Generation}};
    \node[outerBox,fit=(VS)(AVS)] (BVS) at ($(VS) !.6! (AVS)$) {};

    \node[innerBox,right of=VS, xshift=1.25in] (M) {\bf Manipulator};
    \node[detailBox,above of=M]  (AM) {\parbox{1in}{\centering \footnotesize 
        Path Planning}};
    \node[outerBox,fit=(M)(AM),draw] (BM) at ($(M) !.6! (AM)$) {};

    \node[innerBox,above of=VS, yshift=1.25in] (AR) {\bf Meta AR};
    \node[detailBox,below of=AR] (AAR) {\parbox{1in}{\centering \footnotesize
        Meta SDK \\ Dynamic Menu}};
    \node[outerBox,fit=(AR)(AAR),draw] (BAR) at ($(AR) !.6! (AAR)$) {};

    \node[innerBox,right of=AR, xshift=1.25in] (TD) {\bf The TDS};
    \node[detailBox,below of=TD]  (ATD) {\parbox{1in}{\centering \footnotesize
        Intent Selection}};
    \node[outerBox,fit=(TD)(ATD),draw] (BTD) at ($(TD) !.6! (ATD)$) {};

    \node[draw,fit=(BVS)(BM)(BAR)(BTD),very thick, inner sep=6pt,
    minimum height=2.25in, rounded corners=2pt] (HF) at ($(BAR) !.5! (BM)$) {};
    \node[anchor=north] at (HF.south) {System for Hands-Free Robot Assistance};

    \draw[<-, very thick] (BAR.220) -- (BVS.140) node[midway,right] 
        {\parbox{0.75in}{\scriptsize Object class\\Object location}};
    \draw[->, very thick] (BAR.310) -- (BVS.50) node[midway,right] 
        {\scriptsize RGB-D image};
    \draw[->, very thick] (BTD.west) -- (BAR.east);

    \draw[->, very thick](BAR.345) to [out=-10,in=100] (BM.140)
        node[right,anchor=south west,yshift=1ex]
        {\scriptsize Selected action};
    \draw[->, very thick](BVS.east) -- (BM.west) node[below,midway]{\parbox{0.35in}
        {\scriptsize \centering Pose \\ Grasp}};

    \draw[->, very thick](BAR.15) -- ++(0.25in,0in) -- ++(0in,0.30in) 
        -- ++(2.10in,0in) node(Menu)[below,xshift=-1em]{\scriptsize Menu};
    \draw[<-, very thick](BTD.east) -- ++(0.25in,0in) |- ++(0.5in,-0.60in) 
        node(HI)[below,xshift=-1.75em]{\scriptsize Human intent};
    \draw[->, very thick](BM.east) -- ++(0.25in,0in) |- ++(0.7in,-0.4in) 
        node(tM)[below,xshift=-1.75em]{\scriptsize Manipulate};

   \node[anchor=south east,xshift=-0.1in,yshift=0.30in] (DetWB) 
        at (BVS.north west)
        {\includegraphics[width=1.5in]{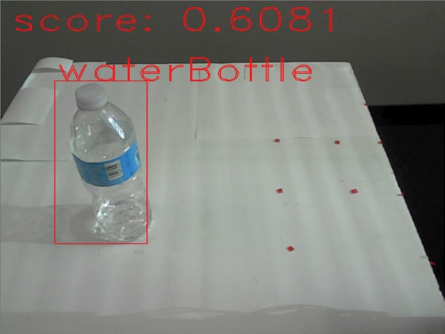}};
   \node[anchor=north] at (DetWB.south){\footnotesize Object Detection};

   \node[anchor=north east,xshift=-0.1in,yshift=0.05in] (bbox) 
        at (BVS.north west)
        {\includegraphics[width=1.5in]{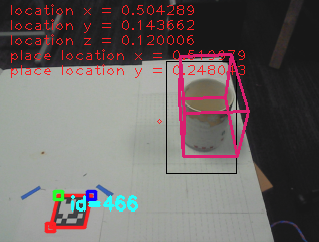}};
   \node[anchor=north] at (bbox.south) {\footnotesize Object bounding box};

   \node[anchor=north west,yshift=0.35in] at (Menu.north east)
        {\includegraphics[height=1in,clip=true,trim=5.4in 0in 0in 0in]
            {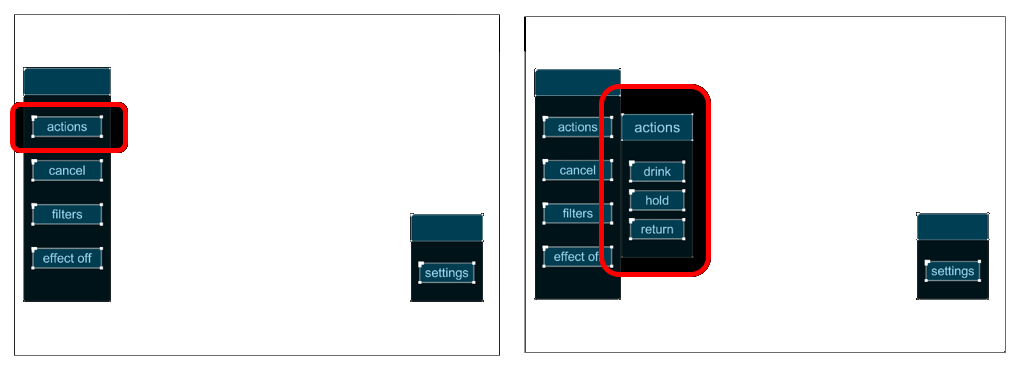}};
   \node[anchor=west] (ImTD) at (HI.east)
        {\includegraphics[width=1.0in]{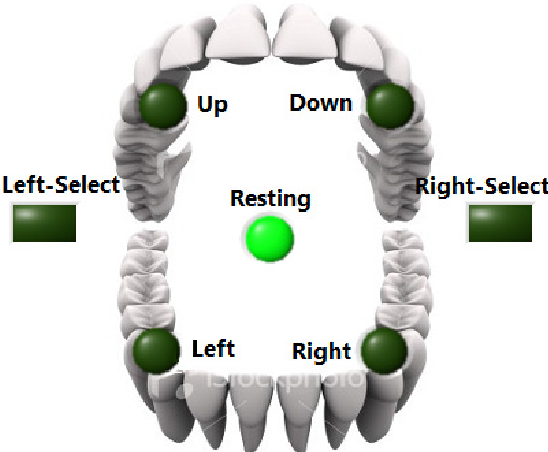}};
   \node[anchor=west] at (tM.east)
        {\includegraphics[height=1in,clip=true,trim=15.75in 2.5in 0in 2in]
            {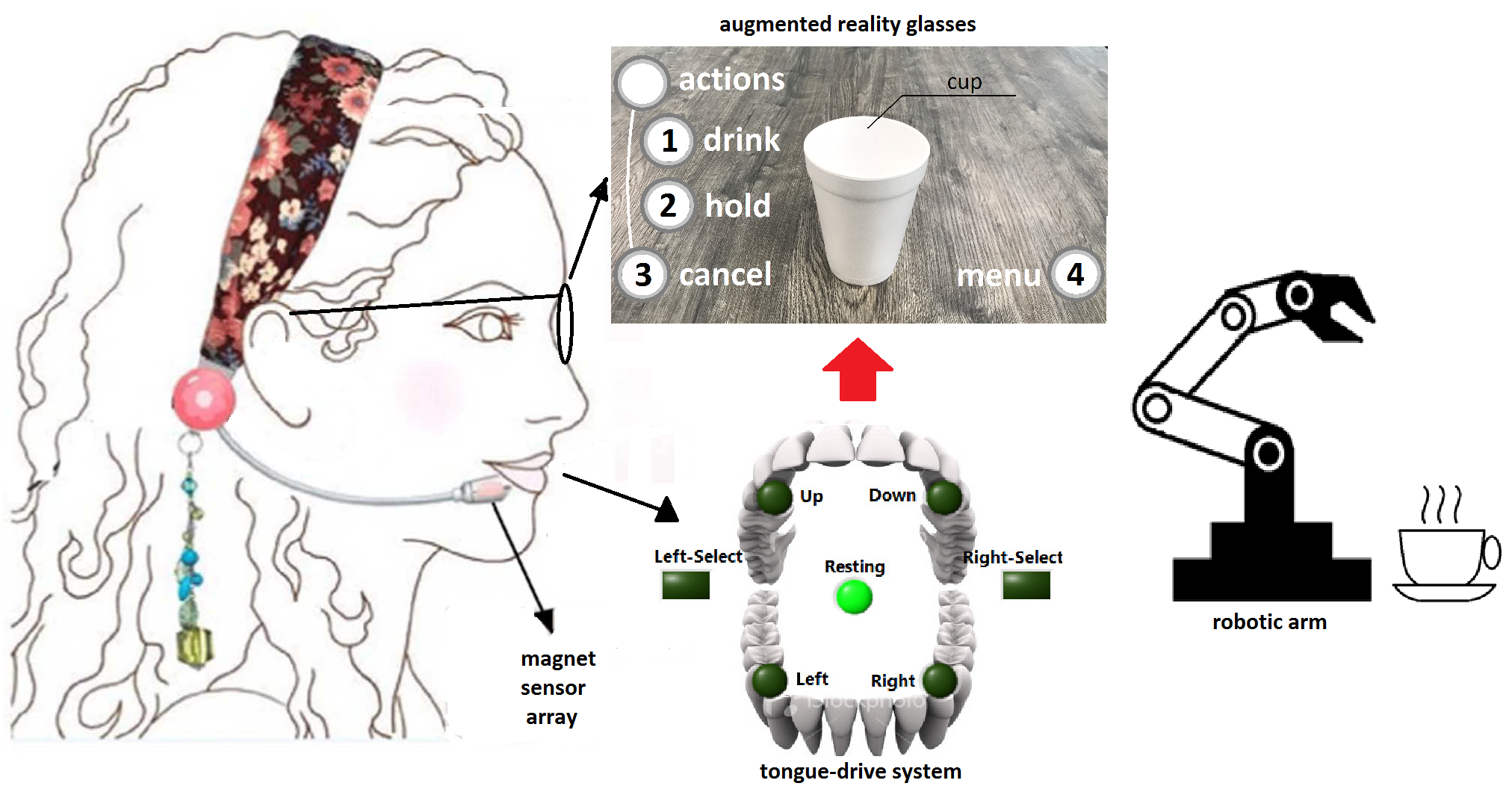}};

  \end{tikzpicture}
  \vspace*{-0.1in}
  \caption{Block diagram of data flow for proposed system modules.
  Top-left: AR glasses recieves RGB-D images; bottom-left: vision system
  performs object detection, localization and grasp detection; up-right:
  TDS receives user's input and triggers robotic arm; bottom-right: a
  7-DOF robotic arm performs manipulation based on human intent.
  \label{system} \label{metamenu} \label{arm}}
  \vspace*{-0.1in}
\end{figure*}
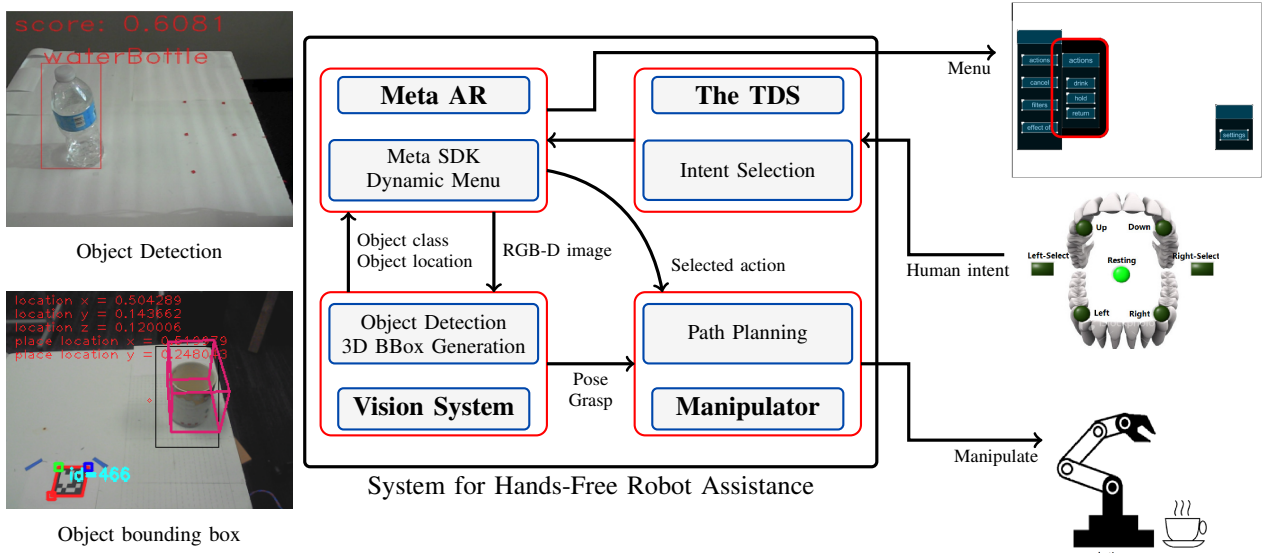

This section describes the human-robot collaborative system,
with Figure \ref{system} depicting the structure of the 
human-in-the-loop system.  There are two main sub-systems:
the autonomous robot (bottom row of blocks) and the human interface (top
row of blocks).   The augmented reality sensors provide visual input to
the {\em Vision System} block consisting of RGB-D images.
After interpretating scene, it generates a corresponding virtual
menu of actions for the manipulator to execute ({\em META AR} block).
Once the AR presents the virtual menu to the user, it waits for the
user's intent as feedback, triggered via {\em the TDS}. 
The TDS input modality enables hands-free operation by mapping tongue
movements to button press operations for virtual menu selection. 
The selected intent will then trigger the manipulator to
autonomously complete tasks ({\em Manipulator} block).
The remainder of this section describes
the augmented reality (\S \ref{ARinterface}),  
vision (\S \ref{visionSys}), and
manipulator (\S \ref{handy}) systems. 

%
%

\subsection{Interface: Egocentric Vision through AR glasses}
\label{ARinterface}

The AR system, a META-1 Developer's Kit, plays the critical
role of transmitting rich visual information between the human and the
autonomous robot sub-systems.  Using AR to visualize actions and provide
context-based menuing systems is more efficient and intuitive 
\cite{LeEtAl_THMS[2014]}, when compared to other modalities.  Further,
the AR system's visual sensors provide a view to the robot similar to
the user's.  The processed scene matches the user's field of view. 





As shown in Fig. \ref{metamenu} (top-right), there is an AR menuing
system for detected objects.  The interface is a Unity3D canvas with 
interactive buttons controlled by the TDS.

\subsection{Vision Interpretation of Users Environment}
\label{visionSys}


\subsubsection{object detection}
The vision system adopts the state-of-the-art deep neural network
architecture, YOLO \cite{redmon2016you}, to recognize objects in a
scene. YOLO is a convolutional neural network with 24 
convolutional layers followed by 2 fully connected layers.  
YOLO's design involves a simpler pipeline and a unified architecture for
improved run-time.  Some YOLO implementations achieve 150 fps processing
rates, which meet real-time requirement for vision-based applications.
To operate with high accuracy for the intended application,
the model is pre-trained on the PASCAL VOC 2007 train/val
$+$ 2012 train/val datasets. Fine-tuning uses a manually collected
dataset of office table objects.  

\subsubsection{object localization}
Manipulation and planning require the object location with respect to the 
manipulator.  To simplify the overall system, the manipulator base is
assumed to be fixed, as well as the surface the objects will lie on.
Establishing the variable AR camera reference frame relative to the fixed
manipulator frame involves localizing the camera using an ARUCO 
\cite{munoz2012aruco} placed on the working surface (Fig. \ref{system}, left).
The 2D bounding box output from the object detection stage is processed
against the calibrated depth image to crop the point cloud region of
interest for post processing. Region growing segmentation \cite{PCL} 
crops the point cloud, with the largest cluster kept (as a denoising step). 
After removing the points belonging to the table surface, the object of
interest remains. From the point cloud, 3D bounding boxes are obtained
for object localization and manipulation purposes.  

\subsubsection{graspable locations}
A second deep neural network architecture recognizes graspable locations
for robotic manipulation. 
Our architecture for grasp detection is described in details in
\cite{1802.00520} with RGB-D input and confidence score output. 
The network is pre-trained on COCO-2014 \cite{lin2014microsoft} and
finetuned on the Cornell dataset \cite{cornell2013} with 1000 augmented
data each. 
This network outputs a list of grasp candidates with a 5D grasp
rectangle representation and corresponding confidence score to inform
the manipulator planning.    
A 5D grasp rectangle representation, $g = \{ x,y, w,h, \theta\}$,
describes grasp configurations for a parallel plate gripper.
The coordinates $(x,y)$ are the center of the rectangle, $\theta$ is the
orientation of the rectangle, and $(w, h)$ are the width and height. 

%

\subsection{Interface: Human Intent to Autonomous Manipulation}
\label{handy}


Once the high-level manipulation command is selected by the user via TDS \cite{huo2008magneto}, all
relevant information for planning is sent to the {\em Manipulator}
system component, whose role is to plan the movement of a 7 degree of
freedom redundant manipulator with a general purpose gripper.
Path planning for manipulation is performed via a modified MoveIt!
package in ROS. The modification admits path planning with mixed initial
and final configurations \cite{Keselman[2014],KeVeVe_ICRA[2014]},
thereby avoiding the need to solve the inverse kinematics of the
redundant manipulator.  
The initial configuration is the current joint state of the manipulator, while 
the final configuration is the desired gripper pose (position and
orientation).
The grasping task relies on the object location and the approaching
direction as estimated by the vision system, which are input to the path
planner model of manipulator. The manipulator autonomously completes the
task without further user input.

%

\section{Experiments and Evaluation}



Evaluation of the assistive system involved a pick and place task. 
The goal is to pick up an object and place it at a user specified location
on the table. Upon starting all processes, the system detects in
real-time objects in the field of view, then waits for the user to
select the object and the action. Selection is triggered by the TDS
based on the AR menu.
Once the pick command is selected, a cross marker is shown at the center of
the field-of-view for user specification of the placement location. The 
marker is projected on the table for 3 dimensional location relative to
robotic arm base for manipulation. After the user triggers the menu again
(place option), the placement is autonomously executed. 

We tested on 10 different types of objects commonly seen.
Each object undergoes 5 trials. We employ the same evaluation criteria
and experimental setup as \cite{LoStLo_ICORR[2017]}. An object is randomly
placed on the visible and reachable area to start the experiment. The
target placement location is 30 cm away from pickup location. 
Placement success means the object is 
within 
a 1cm larger boundary of the
specified location \cite{LoStLo_ICORR[2017]}. 
We compare our semi-autonomous AR+TDS pipeline with manual Cartesian
control, whereby the user controls, via keyboard, the end-effector with
9 commands (rotation, open and close end-effector and 6DOF movement). 

All experiments were carried out with two computers. The vision and manipulator modules are running on Linux machine with Intel core i7-4790K @ 4.00GHz 
and Nvidia Titan-X 
GPU. The TDS and Meta AR modules are implemented on Windows machine with Intel core i5-760 @ 2.80GHz 
due to Windows dependency of APIs. TCP/IP is utilized for communication between machines.       

\begin {table*}[t]
  \centering
  \caption {Comparisons of Cartesian controlled (left number) and 
            TDS+AR controlled (right number) methods on pick-n-place \label{comparison}}
  \small
  \begin{tabular}{ | l | c | c | c | c | c |  }
    \hline
    \bf{Cartesian / TDS+AR} & picked up  & pickup time (s)  & place    &
    place time (s)  & \# commands \\ \hline
    
    {stapler}    
                & 5 / 3 & 69.5\SI{\pm  7.7}\ / 15.7\SI{\pm 0.7}  
                & 5 / 3 & 77.2\SI{\pm  8.9}\ / 12.5\SI{\pm 1.3} & 14.6 / 2 
                \\ \hline
    {spoon}  
                & 5 / 5 & 63.4\SI{\pm  8.7}\ / 16.5\SI{\pm 0.7}  
                & 5 / 5 & 77.7\SI{\pm 19.1}\ / 13.7\SI{\pm 1.9} & 10.4 / 2 
                \\ \hline
    {banana} 
                & 5 / 5 & 67.0\SI{\pm 20.7}\ / 20.9\SI{\pm 3.8} 
                & 5 / 5 & 64.4\SI{\pm 15.4}\ / 14.2\SI{\pm 0.9} & 11.2 / 2
                \\ \hline
    {screw driver} 
                & 5 / 5 & 68.0\SI{\pm  7.7}\ / 20.8\SI{\pm 2.7}  
                & 5 / 5 & 63.6\SI{\pm 13.6}\ / 14.6\SI{\pm 1.5} & 11.4 / 2  
                \\ \hline
    {bowl}   
                & 5 / 4 & 53.6\SI{\pm  2.1}\ / 14.7\SI{\pm 0.7}  
                & 5 / 4 & 84.8\SI{\pm 26.6}\ / 15.8\SI{\pm 1.7} & 9.6 / 2  
                \\ \hline
    {ball}   
                & 5 / 3 & 66.6\SI{\pm  9.9}\ / 18.5\SI{\pm 3.4}  
                & 5 / 3 & 87.3\SI{\pm 16.3}\ / 15.6\SI{\pm 1.7} & 11.6 / 2   
                \\ \hline
    {sunglasses}   
                & 5 / 5 & 62.1\SI{\pm  4.8}\ / 18.3\SI{\pm 5.0}  
                & 5 / 5 & 69.5\SI{\pm 16.3}\ / 11.5\SI{\pm 1.7} & 10.0 / 2    
                \\ \hline
    {pliers} 
                & 5 / 3 & 74.9\SI{\pm  6.6}\ / 16.4\SI{\pm 4.2}  
                & 4 / 3 & 92.4\SI{\pm 24.0}\ / 14.3\SI{\pm 1.9} & 6.5 / 2    
                \\ \hline
    {scissor}      
                & 5 / 3 & 63.8\SI{\pm  7.0}\ / 20.2\SI{\pm 4.7}  
                & 4 / 2 & 69.4\SI{\pm 10.0}\ / 13.7\SI{\pm 2.0}  & 4.2 / 2  
                \\ \hline
    {tape}   
                & 5 / 3 & 60.5\SI{\pm  7.9}\ / 19.7\SI{\pm 4.7}  
                & 5 / 3 & 72.0\SI{\pm 19.2}\ / 15.3\SI{\pm 2.6} & 4.6 / 2   
                \\ \hline \hline
    \bf{average}
                & 5.0 / 3.9 & 65.3\SI{\pm 9.4}\ / 18.3\SI{\pm 3.7}  
                & 4.8 / 3.8 & 76.2\SI{\pm 17.2}\ / 14.5\SI{\pm 1.7} & 9.4 / 2
                \\ \hline
  \end{tabular}

  \vspace*{0.125in}
  \hfill
  {\begin{minipage}[t][1.6in][c]{0.675\linewidth}
  \centering
  \caption{Comparisons with Existing Research \label{pnpCup}}
  \small
  \begin{tabular}{ | l | c | c | c | c | c }
    \hline
    \bf{pickup only / pick-n-place}   & {\centering success rate (\%) }
        & time (s) & \# objects & \# trials\\ \hline
    \bf{\cite{LoStLo_ICORR[2017]}}  & -- / 90 & -- / 56 & 1 & 5 \\ \hline
    \bf{\cite{weisz2017assistive}}  & 82 / -- & 92 / -- & 3 & 5 \\ \hline
    \bf{\cite{struijk2017wireless} w/Actuator}   
                                    & 50 / -- & 70.1 / -- & 1 & 10 \\ \hline
    \bf{\cite{struijk2017wireless} w/Cartesian}  
                                    & 80 / -- & 71.3 / --  & 1 & 10 \\ \hline  
    \hline
    \bf{Ours w/AR+TDS}              & 78 / 76 & 18.3 / 32.7  & 10 & 5\\ \hline

  \end{tabular}
  \end{minipage}}
  \hfill
  {\begin{minipage}[t][1.6in][b]{0.3\linewidth}
  \centering
  \includegraphics[height=1.49in]{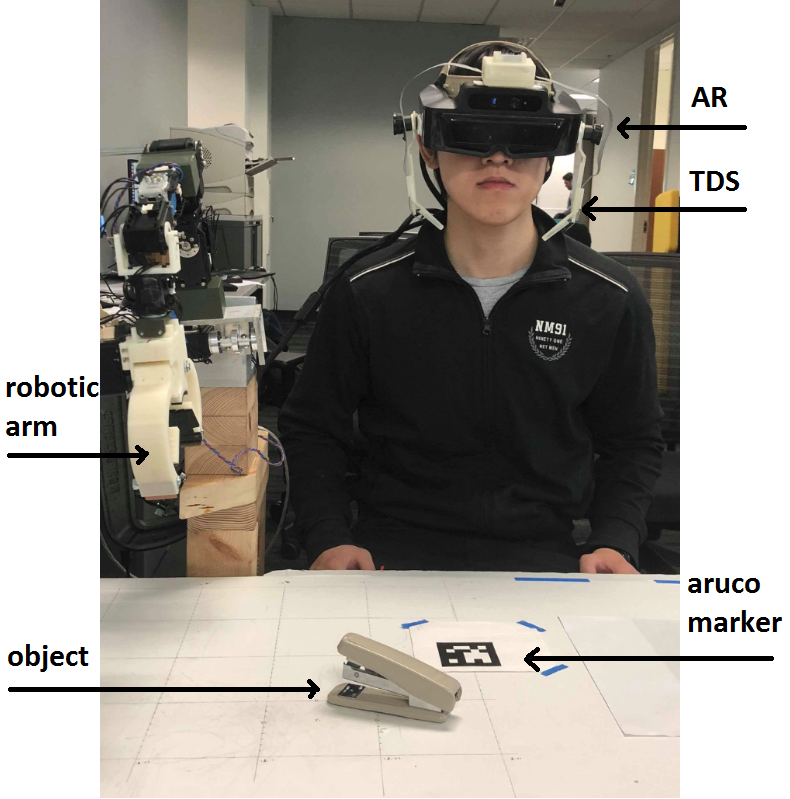}
  \vspace*{-0.1in}
  \captionof{figure}{System setup for experiment.}
  \label{experiment}
  \end{minipage}}
  \hfill
  \vspace*{-0.15in}
\end{table*}

\section{Results}
Evaluation of the system performance involves 
comparisons with manual Cartesian control (Table \ref{comparison}) and
published experiments (Table \ref{pnpCup}).
For the former, we recorded the success rate, average task completion
times, and number of issued commands.  Outcomes and averages for the
five experiments per object are given in Table \ref{comparison}.    
Summarizing robot affordances and automating the task execution reduces the
numer of commands issued by the user. Implicitly this reduction should
lead to a reduced cognitive burden on the user.  The overall operation
speed of the pipeline is 5 times faster then manually controlling the
end-effector in all cases, however the success rate degrades (76\%
versus 96\%).

Comparison to state-of-the-art research in Table~\ref{pnpCup}
provides statistics for two commonly seen manipulation tasks:
pickup and pick-and-place. The table reports the number of objects
tested and trials per object. Note that \cite{wang2017robotic} is 
excluded due to an incompatible test scenario. Compared to
\cite{LoStLo_ICORR[2017]} which moves a bottle with tongue interface 
and a commercial robotic arm in 56s on average, we show ours has
competitive performance with less operation time (32.8s on average).  
The work \cite{struijk2017wireless} applied a tongue operated device to
control a commercial robot end-effector 
step-by-step 
for a pick-up task with 80\% success rate. Our 
system achieves competitive performance with 78\% success rate for a
larger set of objects and with a lower completion time 
(18.11\SI{\pm 2.16} v.s. 70.1\SI{\pm 15.3}s). 
Lastly, \cite{weisz2017assistive} applied non-goggle AR for re-planning
and utilized an EEG to initiate a pickup task. The 
reported average operation time in \cite{weisz2017assistive} is 92s
with 82\% on 3 different objects.  Again, we achieve a comparable
success rate with lower completion time and for a larger set of objects.
Our pipeline achieves a desirable time and accuracy trade-off on wider
variety of objects.

\section{Conclusion}
We presented a collaborative human-robot framework for a person with 
disabilities to guide manipulation tasks. 
Our proposed assistive system provides enhanced autonomy by integrating
vision algorithms with augmented reality and the TDS.  
The human-in-the-loop framework communicates intent and completes tasks
by simplifying the control of manipulation tasks. 
We perform experiments to illustrate the effectiveness of our system
through analysis of the success rate, execution time, and number of
commands issued.
Future studies will include experimental studies with human subjects with upper extremity paralysis
to test the effectiveness and cognitive burden of the proposed system, as well as incorporate
visual servoing algorithms to enhance manipulation performance.



\bibliographystyle{IEEEtran}
\bibliography{IEEEabrv,shorten}

\end{document}